\documentclass[11pt]{article}
\usepackage[a4paper,margin=2.5cm]{geometry}
\usepackage{amssymb}

\usepackage{latexsym}
\usepackage{natbib}
\usepackage[export]{adjustbox}
\usepackage{array}
\usepackage{amsbsy}
\usepackage{footmisc}
\usepackage{amsfonts}
\usepackage{amsthm}
\usepackage{hyperref}
\usepackage{graphicx}
\usepackage{subcaption}
\usepackage{enumerate}
\usepackage{enumitem}
\usepackage{mathtools}
\usepackage{algorithm}% http://ctan.org/pkg/algorithm
\usepackage{algpseudocode}% http://ctan.org/pkg/algorithmicx
\usepackage{color}

\usepackage{booktabs}
\usepackage{multirow}
\usepackage{arydshln}
\usepackage{tikz}
\usepackage{float}
\usepackage{amsmath,amssymb,mathtools}
\usepackage{algorithm}
\usepackage{algpseudocode}   % provides algorithmic environment
\usepackage{xcolor}
\newcommand{\E}{\mathbb{E}}
\newcommand{\Pp}{\mathbb{P}}
\newcommand{\1}{\mathbf{1}}
\newcommand{\base}{\mathrm{base}}
\newcommand{\ft}{\mathrm{ft}}
\usepackage{booktabs}
\usepackage{appendix}

\newtheorem{theorem}{Theorem}[section]

\newtheorem{proposition}[theorem]{Proposition}

\numberwithin{equation}{section}

\title{\textbf{DOHF}: Online Diffusion Fine-tuning with Doob's $h$-transform Guidance}

\author{Zhengyi Guo \and Jiayuan Sheng \and Wenpin Tang \and David D. Yao}

\date{}

\begin{document}
\maketitle

\begin{abstract}
Reward-based diffusion fine-tuning faces practical challenges when desirable outcomes are rare or conditioning corrections are costly to estimate. In this work, we propose \textbf{Diffusion Online $h$-guidance Fine-tuning (DOHF)}, which turns Doob's $h$-transform into a practical online training algorithm. \textbf{DOHF} assigns optimality weights to generated samples, estimates the normalized local correction $\nabla\log h$ under the current rollout policy, and distills it directly into the generative model. Theoretically, we characterize the population-optimal DiffusionNFT update as well as the various classfier free guidance methods through a unified $h$-transform perspective. Methodologically, our framework accommodates black-box and non-differentiable rewards without additional network evaluations. We further show improved alignments under three empirical scenarios. Our work demonstrates how adapting probabilistic conditioning through inexpensive estimation and iterative distillation can improve generative learning across statistical sampling and visual generation.
\end{abstract}

\section{Introduction}

Doob's $h$-transform provides a classical probabilistic mechanism for conditioning a Markov process on a terminal event or reweighting its terminal distribution \citep{doob1957conditional}. 
Given a terminal weight $w(X_0)$, the associated function
$h(x_t,t)=\mathbb{E}\!\left[w(X_0)\mid X_t=x_t\right]$
translates terminal information into a local correction of the intermediate dynamics through $\nabla \log h$.
This perspective has recently become increasingly relevant to diffusion models.
DEFT \citep{denker2024deft} formulates conditional generation through a generalized $h$-transform and learns an auxiliary network.
Related work studies diffusion guidance under hard constraints \citep{guo2026cdg}.
More recently, DOIT \citep{zhu2026DOIT} uses Doob's $h$-transform to construct a training-free adaptation procedure.

From these viewpoints, a central question is how to efficiently estimate local correction for fine-tuning throughout the generative trajectory. The $h$-transform gives an explicit answer.
For a current rollout policy $p^{old}$ and a general reward weight $w$, the desired terminal distribution is
\[
p^{\ft}(x_0)
\propto
w(x_0)p^{old}(x_0),
\]
while its intermediate score differs from that of the current model by exactly $\nabla_{x_t}\log h(x_t,t)$.

Existing $h$-transform approaches use this correction for inference-time guidance \citep{kim2025riskaverse, zhu2026DOIT}, typically relative to a fixed reference model. This can become restrictive when the high-reward region is rare under the original proposal: the trajectories used to estimate the correction may provide little information precisely where adaptation is most needed.
This motivates an online procedure in which the $h$-function itself is defined and estimated under the evolving rollout policy, and the resulting local correction is progressively distilled into the generative model.

In this work, we propose \textbf{Diffusion Online $h$-guidance Fine-tuning (DOHF)}.
At each iteration, samples from the current rollout policy are assigned an optimality weight $w(X_0)$, from which we estimate the corresponding $\nabla\log h$ and distill the resulting correction directly into the generative model.
This online procedure progressively shifts the rollout distribution toward more informative regions, while our simulation-based estimator remains applicable to black-box and non-differentiable rewards.
The same formulation naturally accommodates continuous rewards, hard or soft constraints, and relative optimality objectives.

We further establish a precise connection to DiffusionNFT \citep{zheng2026diffusionNFT}, showing when its population-optimal velocity coincides with the exact $h$-transformed update.
We evaluate \textbf{DOHF} on rare-mode adaptation, constrained contingency-table generation, and SD3.5-Medium text-to-image fine-tuning.
Across these settings, \textbf{DOHF} consistently benefits from jointly adapting the rollout distribution and the $h$-correction, leading to faster adaptation and strong reward alignment when high-reward regions are poorly represented by the original model.

In conclusion, The key of \textbf{DOHF} is not the use of Doob’s \(h\)-transform itself, but turning its normalized local correction into an online, self-improving training target: the correction is re-estimated under the evolving rollout policy and distilled into the model that generates the next round of data.

% ============================================================
\smallskip

\paragraph{Our main contributions are:}

\begin{itemize}
\item \textbf{Optimality as an unified framework.} We formulate reward-based
fine-tuning through a single weight $w:\mathcal{X}\to[0,1]$ interpreted as the
probability of an optimality event $\{O=1\}$. Exponential tilting, hard
constraints $w=\mathbf{1}\{r\ge z\}$, their smooth relaxations, and relative
optimality objectives all become instances of one object, and each induces the
exact target $p^{\mathrm{ft}}\propto w\,p^{\mathrm{old}}$. In contrast to
formulations built on scalar advantages or on a positive/negative split of a
rollout group, this retains the full conditioning structure, and in particular
accommodates indicator-type hard constraints.

\item \textbf{Online $h$-guidance distillation.} We turn the terminal weight
into a local velocity correction $-\tfrac{t}{1-t}\nabla_x\log h$ and regress it
directly into the generative model, re-estimating $h$ under the evolving rollout
policy at every iteration. Our simulation-based estimator reuses the velocity
already computed at the current step, so leads to no additional network evaluations, and $w$ need not be
differentiable. Estimating $h$ and improving the rollout distribution are two
sides of one update: each iteration re-collects trajectories under the tilted
policy, so the correction is estimated where adaptation is most needed.

\item \textbf{An explicit $h$ yields the guidance strength others must tune.}
We provide a population-level characterization of DiffusionNFT,
showing that its update is an $h$-attenuated version of the exact
$h$-transform and becomes exact only under the state-dependent choice
$\beta=2h(x_t,t,c)$. Since \textbf{DOHF}
estimates $h$ explicitly, it realizes this state-dependent strength, whereas
methods that do not must leave $\beta$ a constant --- which attenuates the
update exactly in the low-$h$ regions where high-reward mass is rare. We verify
this prediction and the framework as a whole on rare-mode adaptation,
constrained contingency-table generation, and SD3.5-M text-to-image fine-tuning,
where our formulation is additionally compatible with classifier-free guidance, a design not available to
forward-matching baselines.
\end{itemize}

\paragraph{Related Work.}
Existing $h$-transform methods provide foundations for our work. DEFT \citep{denker2024deft}, hard-constraint guidance \citep{guo2026cdg}, and Doob's Matching \citep{chang2026inferencedoob} learn conditioning corrections, while DOIT \citep{zhu2026DOIT} estimates them through simulation. SVDD \citep{li2024svdd}, twisted SMC \citep{wu2024tds}, and Feynman-Kac steering \citep{singhal2025fk} offer related inference-time approaches. Tilt Matching \citep{potaptchik2025tilt}, Iterative Tilting \citep{pachebat2025iterative}, and Iterative Importance Fine-tuning \citep{denker2026iift} connects tilted distributions with iterative learning.

In parallel, a rapidly growing literature studies reward-based fine-tuning of diffusion models \citep{UZ24, WZ25}.
Policy-based approaches formulate the reverse generation process as a policy and optimize it through RL, including DDPO \citep{black2024ddpo}, FlowGRPO \citep{liu2025flowgrpo}, DanceGRPO \citep{xue2025dancegrpo}, DPOK \citep{fan2023dpok}, and MixGRPO \citep{li2026mixgrpo}. Sampler stochasticity and the resulting training-inference gap are studied in \citep{sheng2025stochasticity}. ReFL \citep{xu2023imagereward} and DRaFT \citep{clark2024} directly differentiate reward signals, while DiffusionDPO
\citep{wallace2023diffusionDPO} learns from preference pairs. Adjoint Matching \citep{domingoenrich2025, guo2026improved} and Score as Action \citep{ZZT24, ZZT24b} cast fine-tuning as stochastic optimal control. Gradient-based construction \citep{zhou2026opsd} and flow-map alignment \citep{holderrieth2026diamond, potaptchik2026metaflow} provide alternatives when reward derivatives are available. Regression-based approaches include Advantage Weighted Matching \citep{xue2026awm} and FlowAWR \citep{fu2026flowawr}, while online reward-weighted flow matching \citep{fan2025online} directly reweights newly generated samples. Reward Score Matching \citep{lee2026rsm}, a path-space perspective \citep{xu2026pathspace}, and reward-based velocity matching \citep{choi2026scaling} clarify the shared structure of these updates.

% ============================================================
\section{Preliminaries}
Throughout, let $c$ denote an optional conditioning context (e.g. a text prompt, a market covariate, or $c=\emptyset$ if unconditional). We suppress $c$ from the notation below whenever it plays no special role.
A pre-trained diffusion/flow model generates
\begin{equation*}
    X_0 \sim p^{\base}(x_0\mid c).
\end{equation*}
Moreover, we generate samples online, so we define the policy for rolling out is $p^{old}(x_0 \mid c)$.

\subsection{A Unified Optimality Weight}
\label{sec:opt_weight}
We define optimality through a single weight function
\[
w:\mathcal X\to[0,1],
\qquad
\E_{p^{old}}[w(X_0)]>0.
\]
We introduce an optimality variable \cite{levine2018rlcp, zheng2026diffusionNFT} defining whether the generated $X_0$ is optimal or not:
\[
O:=\1_{\{U\le w(X_0)\}},
\]
where $U\sim\mathrm{Unif}(0,1)$ is an auxiliary uniform random variable independent of $X_0$.
Intuitively, $X_0$ with higher weight $w(X_0)$ is more likely to be optimal. Then $\Pp(O=1\mid X_0=x_0)=w(x_0)$, and conditioning the base model on $\{O=1\}$ gives the fine-tuned target
\begin{equation*}
p^{\ft}(x_0\mid c):=p^{old}(x_0\mid O=1,c)
=\frac{w(x_0)}{\Pp(O=1\mid c)}\,p^{old}(x_0\mid c)
\;\propto\; w(x_0)\,p^{old}(x_0\mid c).
\end{equation*}
Every choice of $w\in[0,1]$ therefore corresponds to a genuine conditioning event. Note also that $w$ and $Cw$ ($C>0$) induce the same $p^{\ft}$, so any normalization on $w$ has no effect.

\paragraph{Examples.}
(i) \textit{Exponential tilting} \cite{domingoenrich2025}: $w(x_0)=\exp\!\big(\tfrac{r(x_0,c)-r_{\max}(c)}{\tau}\big)$, temperature $\tau>0$.
(ii) \textit{Hard threshold} \cite{guo2026cdg}: $w(x_0)=\1_{\{r(x_0,c)\ge z\}}$, giving $p^{\ft}\propto p^{old}\,\1_{\{r\ge z\}}$.
(iii) \textit{Sigmoid relaxation}: $w(x_0)=\sigma\!\big(\tfrac{r(x_0,c)-z}{\beta}\big)$, $\sigma(u)=(1+e^{-u})^{-1}$. Note that it recovers the hard threshold when $\beta\to0^+$.
(iv) \textit{Loss/statistic-driven weights} \cite{kim2025riskaverse}: $w(x_0)=\psi(\ell(x_0))$ for a nondecreasing $\psi$, e.g. extreme summary statistics of $x_0$ or loss of data $x_0$ defined by upstream models.

\subsection{The {$h$}-Function}
\label{sec:hfunc}

Define $h$-function as the probability of optimality given tuple $(x_t,t,c)$ as
\begin{equation*}
h(x_t,t,c):=\Pp(O=1\mid X_t=x_t,c)=\E\big[w(X_0)\mid X_t=x_t,c\big].
\end{equation*}
In particular $h(x_0,0,c)=w(x_0)$. By Bayes' Rule, the fine-tuned marginal at time $t$ is
\begin{equation}
\label{p_ft}
p_t^{\ft}(x_t\mid c)
=p_t^{old}(x_t\mid O=1,c)
=\frac{1}{Z(c)}\,p_t^{old}(x_t\mid c)\,h(x_t,t,c),
\end{equation}
where $Z(c)$ does not depend on $x_t$. Hence
\begin{equation}\label{score_decomp}
\nabla_{x_t}\log p_t^{\ft}(x_t\mid c)
=\nabla_{x_t}\log p_t^{old}(x_t\mid c)
+\nabla_{x_t}\log h(x_t,t,c),
\end{equation}
so the score correction for fine-tuning is $\nabla_{x_t}\log h(x_t,t,c)=\nabla_{x_t}\log\E\big[w(X_0)\mid X_t=x_t,c\big]$.

% ============================================================
\subsection{Flow Matching}

Consider Gaussian flow matching
\begin{equation*}
    X_t
    =
    (1-t)X_0+tX_1,
    \qquad
    X_1\sim\mathcal N(0,I),
    \qquad
    t\in[0,1],
\end{equation*}

Let $v^{old}(x,t,c)$ denote the velocity field corresponding to old policy. For this Gaussian interpolation, the velocity and score are related (see Appendix~\ref{appx_score_velocity} for derivation) through
\begin{equation*}
v(x,t,c)
=
-\frac{
x+t\nabla_x\log p_t(x\mid c)
}{1-t}.
\end{equation*}
By Equation~\eqref{score_decomp}, the corresponding fine-tuned velocity provides a learning target:
\begin{equation}\label{eq:v_target}
v^{\ft}(x,t,c)
=
v^{old}(x,t,c)
-
\frac{t}{1-t}
\nabla_x\log h(x,t,c).
\end{equation}

% ============================================================
\section{Diffusion
Online $h$-guidance Fine-tuning}

For Flow Matching, the backward process can be reformulated as a SDE \cite{huang2021variational, TZ24tut}:
\begin{equation}
dX_t
=
\left[
(1+\lambda_t^2)
v^{old}(X_t,t,c)
+
\frac{\lambda_t^2}{1-t}X_t
\right]dt
+
\lambda_t
\sqrt{\frac{2t}{1-t}}
\,d\overline B_t,
\label{eq:flow_sde}
\end{equation}
where $\lambda_t$ is a parameter that we can choose, and when $\lambda_t = 0$ it reverts to the original ODE form. 
Consider discrete reverse times $0=t_0<t_1<\cdots<t_L=1$, and let $\phi_\ell^{old}(x_{t_{\ell-1}}\mid x_{t_\ell},c)$ denote the backward Gaussian transition kernel, with mean $\mu_\ell(x_{t_\ell})$ and variance $\sigma_\ell^2 I$.

In order to do the regression~\eqref{eq:v_target}, we need $\nabla_x\log h(x,t,c)$. Fortunately, we could use Monte Carlo Method to estimate it during the online fine-tuning process.

\subsection{Simulation-Based Estimator of $\nabla_x\log h(x,t,c)$ \cite{zhu2026DOIT}}\label{sec:MC}

Given $X_{t_\ell}=x_{t_\ell}$, simulate $M$ backward trajectories $\{x_{t_{\ell-1}}^{(m)},\ldots,x_0^{(m)}\}_{m=1}^M$ using $\phi^{old}$. Define
\begin{equation*}
W_m := w\big(x_0^{(m)}\big),
\qquad
G_m := \nabla_{x_{t_\ell}}\log\phi_\ell^{old}\big(x_{t_{\ell-1}}^{(m)}\mid x_{t_\ell},c\big)
=\frac{1}{\sigma_\ell^2}\big[\nabla_{x_{t_\ell}}\mu_\ell(x_{t_\ell})\big]^\top\big(x_{t_{\ell-1}}^{(m)}-\mu_\ell(x_{t_\ell})\big).
\end{equation*}
The plug-in estimator is
\begin{equation}\label{est:MC}
\widehat{\nabla\log h}_{\mathrm A}
=\frac{\sum_{m=1}^M W_m G_m}{\big(\sum_{m=1}^M W_m\big)\vee \tau_\ell},
\end{equation}
with small truncation level $\tau_\ell>0$ for numerical stability. To alleviate the cost by $M$ rollouts of length $\ell$, we use  one-step-lookahead surrogate
\begin{equation*}
\hat x_0^{(m)} = x_{t_{\ell-1}}^{(m)} - t_{\ell-1}\,v^{old}(x_{t_\ell},t_\ell,c),
\qquad
x_{t_{\ell-1}}^{(m)}\sim\mathcal N\big(\mu_\ell(x_{t_\ell}),\sigma_\ell^2 I\big),
\end{equation*}
in place of $x_0^{(m)}$, at zero extra NFE (reusing $v^{old}(x_{t_\ell},t_\ell,c)$ already computed). Moreover, it does not require $w$ to be differentiable, which is more general for various terminal rewards mentioned in Section~\ref{sec:opt_weight}. If $w$ is differentiable, we also provide an alternative estimator using finite difference method, which is provided in Appendix~\ref{appdx:alter_est}.

\subsection{Training and Sampling}
 We instantiate the training \ref{alg:h-distill-train} for Text-to-Image fine-tuning, and for generalized task, prompt could be omitted. In terms of sample rolling-out, we use a slow-mix strategy: starts from on-policy and smooth it with exponential moving average. Note that we index time so that t=1 is pure noise and generation runs from $t_L=1$ to $t_0=0$.

For sampling process, especially for Text-to-Image task, we could also implement the similar way used in Classifier-free Guidance \cite{ho2022cfg}, as we set some proportion of the prompt to be empty in  training process, in order to improve its unconditional generation. For other tasks, just use fine-tuned $v_\theta$ to sample. For generalization purpose, we provide Algorithm \ref{alg:h-distill-sample} used in T2I task.

\begin{algorithm}[t]\label{algo:dohf}
\caption{\textbf{D}iffusion \textbf{O}nline \textbf{h}-guidance \textbf{F}ine-tuning (\textbf{DOHF})}
\label{alg:h-distill-train}
\begin{algorithmic}[1]
\Require Pretrained $v^{\text{base}}$, prompt set $\mathcal{C}$, reward model $r$, schedule $\{t_i\}$, slow-mix schedule $\eta_k$
\State Initialize model weight $v_\theta\leftarrow v^{\base}$, and sample-generation policy $v_\phi\leftarrow v_\theta$ 
\For{$k=1,2,\ldots$}
  \For{prompt $c$ in mini-batch}
    \State Sample $t_\ell$ uniformly; roll out $v_\phi(\cdot,\cdot,c)$ to obtain $x_{t_\ell}$
    \State Compute $\widehat{\nabla\log h}(x_{t_\ell},t_\ell,c)$ \Comment{via \ref{est:MC} or \ref{est:fd} }
    \State $v^{target} \leftarrow \mathbf{stopgradient}\!\left(v_\phi(x_{t_\ell},t_\ell,c)-\tfrac{t_\ell}{1-t_\ell}\,\widehat{\nabla\log h}\right)$
    \State $\mathcal L \leftarrow \mathcal L + \|v_\theta(x_{t_\ell},t_\ell,c)-v^{target}\|_2^2$
  \EndFor
  \State Update $\theta$ by one step on $\nabla_\theta\mathcal L$
  \State $v_\phi\leftarrow \eta_k v_\phi+(1-\eta_k)v_\theta$ \Comment{slow-mix sample-generation policy}
\EndFor

\Return $v_\theta$
\end{algorithmic}
\end{algorithm}

\begin{algorithm}[t]
\caption{Sampling with Classifier-Free Guidance}
\label{alg:h-distill-sample}
\begin{algorithmic}[1]
\Require Trained $v_\theta$, prompt $c$, guidance scale $\gamma\!\ge\!0$,
  schedule $\{t_i\}_{i=0}^{N}$
\State $x \leftarrow z \sim \mathcal{N}(0, I)$
\For{$i = 0, 1, \ldots, N-1$}
  \State $v_{\text{cond}}   \leftarrow v_\theta(x,t_i, c)$, \quad $v_{\text{uncond}} \leftarrow v_\theta(x,t_i, \varnothing)$
  \State $v \leftarrow v_{\text{uncond}} + \gamma\,(v_{\text{cond}} - v_{\text{uncond}})$
  \State $x \leftarrow x + v\,(t_{i+1} - t_i)$
\EndFor

\Return $x$
\end{algorithmic}
\end{algorithm}

\subsection{Population-Level Relation to DiffusionNFT}
\label{sec:nft_connection}

DiffusionNFT~\citep{zheng2026diffusionNFT} is the closest online
fine-tuning method to \textbf{DOHF}. Although its positive--negative construction
does not explicitly estimate an $h$-transform, its population optimum
admits an exact characterization in terms of the same $h$-function.

Let
\[
h(x_t,t,c)
=
\mathbb{E}_{p^{\mathrm{old}}}
\left[w(X_0)\mid X_t=x_t,c\right].
\]
DiffusionNFT parameterizes its positive and negative branches as
\[
v_\theta^{+}
=
(1-\beta)v^{\mathrm{old}}+\beta v_\theta,
\qquad
v_\theta^{-}
=
(1+\beta)v^{\mathrm{old}}-\beta v_\theta,
\]
where $\beta>0$ is a fixed guidance coefficient.

\begin{proposition}
At the population optimum of the double-branch DiffusionNFT objective,
the learned velocity satisfies
\[
v_\theta^{*}(x_t,t,c)
=
v^{\mathrm{old}}(x_t,t,c)
-
\frac{2h(x_t,t,c)}{\beta}
\frac{t}{1-t}
\nabla_{x_t}\log h(x_t,t,c).
\]
Consequently, its population optimum coincides with the exact
$h$-transformed velocity
\[
v^{\mathrm{ft}}
=
v^{\mathrm{old}}
-
\frac{t}{1-t}\nabla_{x_t}\log h
\]
if and only if
\[
\beta=2h(x_t,t,c).
\]
\end{proposition}
A complete derivation is provided in Appendix~\ref{appdx_relation_nft}.

This result reveals a key distinction between the two methods.
With a fixed $\beta$, DiffusionNFT effectively follows the unnormalized
direction $\nabla h=h\nabla\log h$; hence, its update is attenuated in
low-$h$ regions where optimal samples are rare. \textbf{DOHF} instead directly
estimates the normalized correction $\nabla\log h$ under the evolving
rollout policy and distills it into the model. It therefore realizes the
exact $h$-transformed update without requiring a state-dependent guidance
hyperparameter.

% ============================================================

\section{Experiment}
\subsection{Gaussian Mixture}
To isolate the effect of online iteration from confounders, we conduct a controlled study on a 2D synthetic problem. The base
distribution is an imbalanced two-mode Gaussian mixture
\[
p_{\mathrm{data}}(x) \;=\; 0.01 \cdot \mathcal{N}\!\big(x;\, \mu_0,\, \Sigma\big)
\;+\; 0.99 \cdot \mathcal{N}\!\big(x;\, \mu_1,\, \Sigma\big),
\]
with $\mu_0 = (-1, -1)^\top$, $\mu_1 = (1, 1)^\top$, and $\Sigma = 0.16\,\mathbf{I}$.
A rectified-flow velocity network (a small MLP) is first pretrained on
samples from $p_{\mathrm{data}}$.

The reward is the \emph{soft posterior probability of belonging to the
minority mode}:
\[
r(x) \;=\; \mathbb{P}(\text{cluster} = 0 \,|\, x)
       \;=\; \frac{0.01 \cdot \mathcal{N}(x;\, \mu_0,\, \Sigma)}
                   {0.01 \cdot \mathcal{N}\!\big(x;\, \mu_0,\, \Sigma\big)
\;+\; 0.99 \cdot \mathcal{N}\!\big(x;\, \mu_1,\, \Sigma\big)}
       \;\in\; (0, 1).
\]
This turns the alignment task into an extreme mode-seeking problem:
the fine-tuned model should transport probability mass from the majority
mode $\mu_1$ into the minority mode $\mu_0$.

We compare two reward-tilting strategies, both starting from the same
pretrained base:
\begin{itemize}
  \item \textbf{Off-policy $h$-guidance \cite{guo2026cdg}.} A separate value network
    $h_\phi(x_t, t) \approx \mathbb{E}[r(X_0)\,|\,X_t{=}x_t]$ is regressed
    on rollouts drawn \emph{exclusively} from the frozen base, then applied
    as guidance at inference via $\hat v = v_{\text{base}} - \tfrac{t}{1-t}
    \nabla \log h_\phi$.
  \item \textbf{DOHF(ours).} The same
    $h$-transform correction is instead distilled into the velocity
    network itself via online data collection.
\end{itemize}

\begin{figure}
    \centering
    \includegraphics[width=0.9\linewidth]{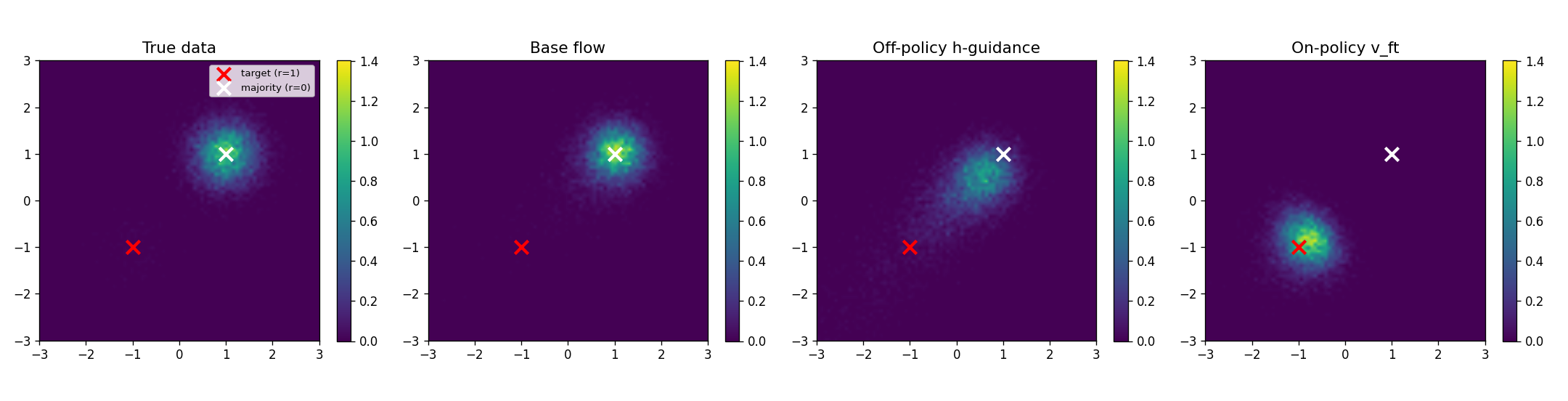}
    \caption{Visualization of on/off-policy h-transform guidance.}
    \label{fig:gaussian_comparison}
\end{figure}

\begin{table}[t]
\centering
\small
\begin{tabular}{lccc}
\toprule
Method & Iterations & Minority-mode fraction & Mean $\pm$ std \\
\midrule
True data                       & --   & \phantom{00}0.96\% & $(0.98, 0.98) \pm  (0.44, 0.44)$ \\
Base pretrained                             & --   & \phantom{00}1.12\% & $(0.99, 0.98) \pm  (0.42, 0.41)$ \\
Off-policy $h$-guidance                     & 2000 & \phantom{00}14.88\% & $(0.29, 0.25) \pm  (0.73, 0.71)$ \\
\textbf{DOHF(ours)}  & \phantom{0}200 & \textbf{100.00\%} & $(-0.82, -0.88) \pm  (0.38, 0.38)$ \\
\bottomrule
\end{tabular}
\caption{Fraction of samples falling in the minority cluster after fine-tuning.}
\label{tab:synthetic_2d}
\end{table}

Table~\ref{tab:synthetic_2d} summarizes the outcome. Off-policy guidance takes $2000$ update steps and still lifts the
minority-mode fraction to some extent ($1.12\% \to
14.88\%$): most of its training rollouts still land in the majority mode. In contrast, online $h$-distillation reaches
$100\%$ minority-mode concentration after just $200$ steps
--- one order of magnitude fewer updates for a much better outcome: better preserves the target distribution with closer mean $(-0.82, -0.88)$ and standard deviation  $(0.38,0.38)$. Because each iteration re-collects trajectories under
tilted policy, the reward-relevant regions of state space are
explored progressively as the model itself improves. For visualization, see Figure~\ref{fig:gaussian_comparison}.

\subsection{Contingency Table}
\label{sec:contigency_table}
We consider contingency-table generation as an application of endogenous conditional sampling under the framework of \cite{guo2026cdg}. Classical references include \cite{DiaconisEfron1985, DiaconisAAS}, with connections to Burnside processes and double cosets discussed in \cite[Section 3]{Diaconis2025}. Such tables also arise in financial-network and systemic-risk applications \cite{albanesi2019, greenwood2022}.

\paragraph{Objective.}
Let $X=(x_{ij})\in\mathbb{Z}_{+}^{m\times n}$ be contingency table with prescribed row and column sums
\begin{equation}\label{eq:rc}
    \sum_j x_{ij}=r_i,
    \qquad
    \sum_i x_{ij}=c_j.
\end{equation}
In practice, these marginals may be noisy or partially observed, so we treat them as soft structural targets.

Given a pretrained flow model $v^{\base}$, define
\[
\mathcal{S}_2
=
\sum_i\left(\sum_j x_{ij}-r_i\right)^2
+
\sum_j\left(\sum_i x_{ij}-c_j\right)^2.
\]
Exact feasibility corresponds to $\mathcal{S}_2=0$. Since exact matches are extremely rare for large tables, directly using the indicator reward leads to high-variance $h$-guidance estimates, usually there is no such table in a batch satisfying $\mathcal{S}_2=0$. We therefore use the smooth reward
\[
w(x)=\exp(-\beta \mathcal{S}_2),
\]
where larger $\beta$ more closely approximates the hard constraint.

\paragraph{Birthday--deathday contingency table.}
We evaluate on the classical $12\times12$ birthday--deathday table of 82 descendants of Queen Victoria \cite{diaconis1998algebraic}. The target margins are
\[
r=(6,5,5,12,12,3,10,7,3,7,9,3),
\qquad
c=(13,4,7,10,8,4,5,3,4,9,7,8),
\]
with total count $82$, the original table is provided as Table~\ref{tab:birthday_deathday}.

We pre-train a flow model that generates nonnegative integer-valued
$12\times12$ tables with total count $82$; implementation details are given in Appendix~\ref{contingency_pretrained}. Starting from the same pretrained model, we compare frozen off-policy $h$-guidance with \textbf{DOHF} using the reward $\exp(-\beta\mathcal{S}_2)$, where we take $\beta=0.02$.

\begin{table}[t]
\centering
\begin{tabular}{lcccc}
\toprule
Method 
& Updates 
& Avg. Margin Error $\downarrow$
& Max Margin Error $\downarrow$
& $W_2$ vs.\ DG sampler~\cite{diaconis1995rectangular} $\downarrow$ \\
\midrule
Uniform
& --  & 0.802 & 1.975 & 13.48 \\
Off-policy
& 800 & 0.379 & 1.005 & 11.00 \\
\textbf{DOHF}
& \textbf{300} & \textbf{0.210} & \textbf{0.531} & \textbf{9.84} \\
\bottomrule
\end{tabular}
\caption{Comparison of margin matching and distributional similarity. 
The margin errors are standardized by the corresponding generated marginal standard deviations.}
\label{tab:margin_w2_comparison}
\end{table}

Among 500,000 uniformly sampled pre-training tables, none exactly satisfies all target margins, illustrating the sparsity of the target event and the necessity of using soft targets. As shown in Table~\ref{tab:margin_w2_comparison}, uniform sampling (as how we generate pre-training data) yields an average standardized margin error of 0.802 and a maximum error of 1.975. After only 300 updates, \textbf{DOHF} reduces these errors to 0.210 and 0.531, respectively, outperforming the off-policy baseline, which reaches 0.379 and 1.005 after 800 updates. 

The Diaconis–Gangolli (DG) sampler \cite{diaconis1995rectangular} is a Markov Chain based method, generating approximately uniform contingency tables with fixed margins through repeated local \(2\times2\) integer-preserving moves, which requires initial burn-in (in our experiment 200,000 steps are used). Although we are generating tables with perturbed margins, it is still worthwhile to check the difference between generated distributions. It is shown that \textbf{DOHF} also achieves the smallest Wasserstein distance to the DG sampler baseline among the three methods.

The advantage comes from adapting the rollout distribution toward the target region, which provides increasingly informative samples for subsequent $h$-transform updates. For this experiment, however, we set $v_\phi=v^{\base}$ at each iteration rather than using the EMA update in Algorithm~\ref{alg:h-distill-train}, since the latter empirically caused collapse to nearly identical tables. This modification preserves rollout diversity while retaining the benefit of online $h$-distillation.

\subsection{Image Fine-tuning}
Our flow-based Text-to-Image model fine-tuning task is based on \textbf{SD3.5-Medium} \cite{esser2024} at $256 \times 256$ resolution. We demonstrate the potential of our method trained with the sum of three reward models: \textsc{HPSv2.1} \cite{wu2023humanpreferencescorev2}, \textsc{PickScore} \cite{kirstain2023pickapic} and \textsc{ClipScore} \cite{radford2021}. 

\begin{figure*}[t]
    \centering
    \setlength{\tabcolsep}{10pt}

    \begin{tabular}{c c c c}
        \textbf{SD3.5M w/ CFG}
        & \textbf{DOHF w/o CFG}
        & \textbf{DOHF w/ CFG}
        & \textbf{FlowGRPO} \\[2mm]

        \includegraphics[width=0.18\textwidth]{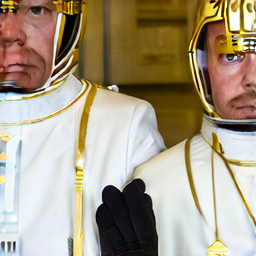}
        & \includegraphics[width=0.18\textwidth]{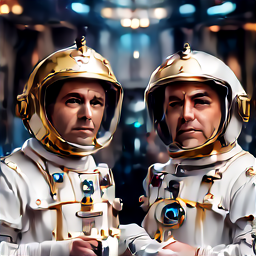}
        & \includegraphics[width=0.18\textwidth]{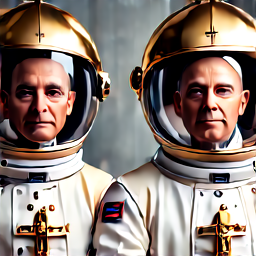}
        & \includegraphics[width=0.18\textwidth]{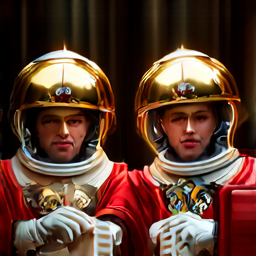} \\

        \multicolumn{4}{c}{
        \small \textit{``Twin obese archbishops in gold space-suits, Vatican launchpad, Hasselblad style."}
        } \\[3mm]

        \includegraphics[width=0.18\textwidth]{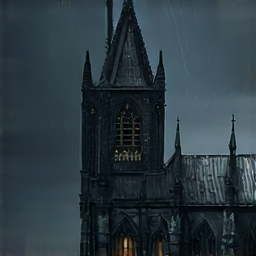}
        & \includegraphics[width=0.18\textwidth]{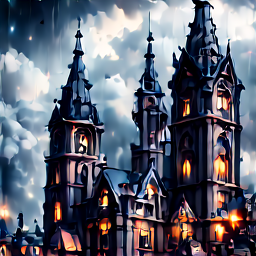}
        & \includegraphics[width=0.18\textwidth]{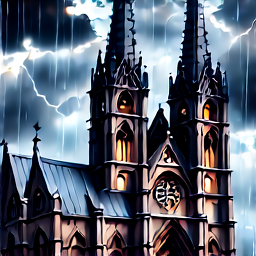}
        & \includegraphics[width=0.18\textwidth]{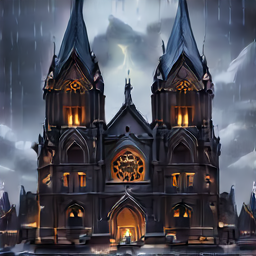} \\

        \multicolumn{4}{c}{
        \small \textit{``Realistic Gothic cathedral in a stormy night."}
        } \\[3mm]

        \includegraphics[width=0.18\textwidth]{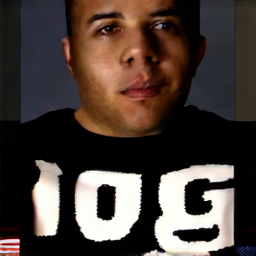}
        & \includegraphics[width=0.18\textwidth]{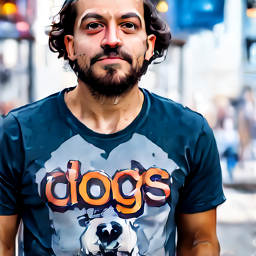}
        & \includegraphics[width=0.18\textwidth]{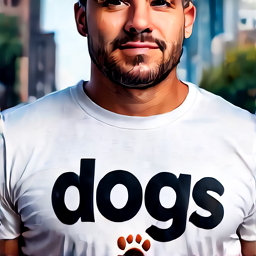}
        & \includegraphics[width=0.18\textwidth]{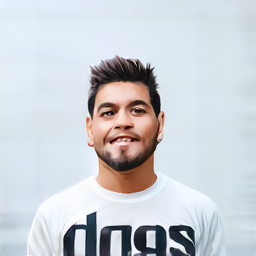} \\

        \multicolumn{4}{c}{
        \small \textit{``A man wearing a shirt that says ``dogs''."}
        }
    \end{tabular}

    \caption{Qualitative comparison across different guidance and fine-tuning methods. The prompts are taken from \textsc{Pick-a-Pic} \cite{kirstain2023pickapic}.}
    \label{fig:qualitative_comparison}
\end{figure*}

We will evaluate our method and other baselines on both in-domain and out-of-domain metrics: \textsc{ImageReward} \cite{xu2023imagereward}, \textsc{Aesthetic Score} \cite{schuhmann2022}, \textsc{GenEval} \cite{ghosh2023geneval} and pair-wise \textsc{LPIPS} \cite{zhang2018lpips}. Note that we used average pair-wise LPIPS per prompt to measure image diversity. We aim to show that rolling out online is not only more efficient compared to off-policy fashion, but also provide a higher reward, combined with better semantic alignment. For Qualitative comparison, refer to Figure~\ref{fig:qualitative_comparison}.

The training and testing prompt set is the ones used in Pick-a-Pic \cite{kirstain2023pickapic}. We finetuned with LoRA($\alpha = 32$, $r=32$). Each iteration consists of $28$ prompts, and generates $M=8$ images per prompt via one-step-lookahead surrogate discussed in Section~\ref{sec:MC}. For both training and testing roll-outs, 40 time steps are applied. Additional details are provided in 
Appendix~\ref{appdx_t2i_exp}.

As shown in Table~\ref{tab:evaluation}, our \textbf{DOHF} methodology works very well even without classifier-free-guidance (CFG) \cite{ho2022cfg} using only 2k iterations, surpassing FlowGRPO \cite{liu2025flowgrpo} on most of the metrics using 9k iterations. Moreover, it performs slightly better than DiffusionNFT \cite{zheng2026diffusionNFT} overall in 2k iterations. To better explain their relationship, detailed discussion are provided in Appendix~\ref{appdx_relation_nft}. 

However, our online $h$-guidance has CFG design which is not available in DiffusionNFT. During training, we set $10\%$ of the prompts to be empty, in order to enforce model to have better unconditional generation, thus $v_{\theta}(\cdot,\cdot,c) - v_{\theta}(\cdot,\cdot,\emptyset)$ acts like guidance. During inference, velocity field is $v_{\theta}(\cdot,\cdot,c) + \gamma \big( v_{\theta}(\cdot,\cdot,c) - v_{\theta}(\cdot,\cdot,\emptyset) \big)$, where $\gamma = 4$ is the default guidance scale. It is shown in Table~\ref{tab:evaluation} that with guidance, all the quality scores increase to some extent, at little cost of diversity. It is worth noting that \textsc{GenEval}, the out-of-domain rule-based reward measuring how accurately generated images satisfy compositional prompts, increase significantly from $44.27\%$ to $62.52\%$, showing its potential in semantic alignment.

\begin{table}[t]
\centering
\resizebox{\textwidth}{!}{
\begin{tabular}{lcccccccc}
\toprule
\textbf{Model} 
& \textbf{\#Iter} 
& \textbf{GenEval} 
& \textbf{PickScore} 
& \textbf{ClipScore} 
& \textbf{HPSv2.1} 
& \textbf{Aesthetic} 
& \textbf{ImgRwd} 
& \textbf{LPIPS} \\
\midrule

SD3.5-M w/o CFG
& -- & 4.49\% & 0.179 & 0.178 & 0.150 & 4.513 & -1.809 & 0.771 \\

SD3.5-M w/ CFG
& -- & 41.35\% & 0.203 & 0.266 & 0.248 & 5.428 & 0.250 & 0.674 \\

SD3.5-L w/ CFG
& -- & 25.00\% & 0.193 & 0.243 & 0.208 & 4.919 & -0.452 & 0.733 \\

\midrule

\textbf{DOHF} w/o CFG
& 2k & 44.27\% & \underline{0.216} & \underline{0.264} & \underline{0.327} & \textbf{6.327} & 1.115 & 0.538 \\

\textbf{DOHF} w/ CFG
& 2k & \textbf{62.52\%} & \textbf{0.218} & \textbf{0.277} & \textbf{0.336} & \underline{6.240} & \textbf{1.329} & 0.521 \\

Off-policy Guidance
& 2k & 19.52\% & 0.184 & 0.201 & 0.177 & 4.739 & -1.320 & \textbf{0.746} \\
& 8k & 38.96\% & 0.197 & 0.240 & 0.249 & 5.422 & 0.190 & \underline{0.655} \\

DiffusionNFT
& 2k & \underline{49.39\%} & 0.214 & 0.261 & 0.305 & 6.151 & \underline{1.118} & 0.525 \\

FlowGRPO
& 2k & 35.89\% & 0.205 & 0.252 & 0.272 & 5.924 & 0.636 & 0.602 \\
& 9k & 45.66\% & 0.211 & 0.259 & 0.302 & 6.162 & 1.020 & 0.588 \\

\bottomrule
\end{tabular}
}
\caption{Evaluation Results under resolution $256 \times 256$. \textbf{Bold}: best; \underline{Underline}: second best.}
\label{tab:evaluation}
\end{table}

\section{Conclusion}

We introduced \textbf{DOHF}, an online diffusion fine-tuning framework based on Doob's $h$-transform. Our method turns reward-based adaptation into a local score or velocity correction that can be estimated online and directly distilled into the generative model. This perspective provides a unified treatment of continuous rewards, hard and soft constraints, and relative optimality objectives, while also clarifying the connection to existing methods such as DiffusionNFT.

Across rare-mode adaptation, constrained contingency table generation, and text-to-image fine-tuning, \textbf{DOHF} consistently demonstrates the benefit of adapting the rollout distribution together with the model, particularly when reward-relevant regions are difficult to sample via base model. These results suggest that online $h$-transform fine-tuning offers a promising alternative to both inference-time guidance and trajectory-based reinforcement learning. More broadly, the framework provides a flexible interface between probabilistic conditioning and generative-model optimization, and may be useful for a wide range of problems involving rare events, structured constraints, black-box rewards, and distributional adaptation.

\clearpage
\bibliographystyle{abbrv}
\bibliography{reference}

\newpage
\appendix
\section{Relation between score and velocity}
\label{appx_score_velocity}
Consider the rectified-flow interpolation
\[
X_t = (1-t)X_0 + t\epsilon,
\qquad
\epsilon \sim \mathcal{N}(0,I).
\]
The corresponding velocity field is
\[
v(x,t)
=
\mathbb{E}\!\left[\epsilon - X_0 \mid X_t = x\right].
\]

Conditioned on $X_0$, we have
\[
X_t \mid X_0
\sim
\mathcal{N}\!\left((1-t)X_0,t^2 I\right),
\]
and hence
\[
\nabla_x \log p_t(x\mid X_0)
=
-\frac{x-(1-t)X_0}{t^2}
=
-\frac{\epsilon}{t}.
\]

Using the score marginalization identity,
\[
\nabla_x \log p_t(x)
=
\mathbb{E}\!\left[
\nabla_x \log p_t(x\mid X_0)
\mid X_t=x
\right],
\]
we obtain
\[
\nabla_x \log p_t(x)
=
-\frac{1}{t}
\mathbb{E}\!\left[\epsilon \mid X_t=x\right].
\]

Therefore,
\[
\mathbb{E}\!\left[\epsilon \mid X_t=x\right]
=
x + (1-t)
\mathbb{E}\!\left[\epsilon-X_0 \mid X_t=x\right]
=
x + (1-t)v(x,t).
\]

Substituting this into the previous expression gives
\[
\nabla_x \log p_t(x)
=
-\frac{x+(1-t)v(x,t)}{t}.
\]
Therefore, we have
\[
v(x,t)
=
-\frac{
x+t\nabla_x\log p_t(x)
}{1-t}.
\]

\section{Alternative Estimator of $\nabla_x \log h$ \cite{kim2025riskaverse}:}
\label{appdx:alter_est}

Approximate $h(x,t,c)\approx w(\hat x_0(x,t,c))$ via the Tweedie point estimate
\begin{equation*}
\hat x_0(x,t,c) = x - t\,v^{old}(x,t,c),
\end{equation*}
so that $\nabla_x\log h(x,t,c)\approx \big[\partial_x \hat x_0(x,t,c)\big]^\top u$, with $u:=\nabla_{\hat x_0}\log w(\hat x_0(x,t,c))$. Moreover, we should avoid the Jacobian $\partial_x\hat x_0$ via a finite-difference step of size $\epsilon>0$:
\begin{equation}\label{est:fd}
\widehat{\nabla\log h}_{\mathrm B}
=\frac{1}{1-t}\left[u+\frac{t^2}{\epsilon}\Big(\nabla_x\log p_t^{old}(x+\epsilon u\mid c)-\nabla_x\log p_t^{old}(x\mid c)\Big)\right],
\end{equation}
where $\nabla_x\log p_t^{old}(x\mid c)=-\frac{x+(1-t)v^{old}(x,t,c)}{t}$. This method doesn't require $M$ rollouts in Estimator~\ref{est:MC}, only 2 evaluations of $v^{old}$ plus one backward pass through $w$. However, it requires $w$ to be differentiable.

\section{Relation to Diffusion Negative-aware Fine-Tuning (NFT)}
\label{appdx_relation_nft}

DiffusionNFT\cite{zheng2026diffusionNFT} arrives at a guidance mechanism built from the optimality variable $r(x_0,c)=p(O{=}1\mid x_0,c)$, which is exactly our weight $w(x_0)$. We show that its guidance direction, and both its single-branch and double-branch training objectives, are special cases of the $h$-transform framework developed above.

\subsection{Distribution split coefficient $\alpha(x_t)$ is equvalent to $h(x_t,t,c)$}

DiffusionNFT defines $\pi^+(x_0\mid c):=\pi^{old}(x_0\mid O{=}1,c)$ and $\pi^-(x_0\mid c):=\pi^{old}(x_0\mid O{=}0,c)$.
It can be shown that the posterior distribution of old policy can be seperated as
\[
\pi^{old}_{0 \mid t}(x_0 \mid x_t,c) = \alpha(x_t)\pi^{+}_{0 \mid t}(x_0 \mid x_t,c) + [1-\alpha(x_t)]\pi^{-}_{0 \mid t}(x_0 \mid x_t,c),
\]
where
\[
\alpha(x_t):=\frac{\pi_t^+(x_t\mid c)}{\pi_t^{old}(x_t\mid c)}\cdot\E_{\pi^{old}(x_0\mid c)}[w(x_0)].
\]
Also,
\begin{equation}
\label{eq:velocity_split}
    v^{old}(x_t, t, c) = \alpha(x_t)v^{+}(x_t, t, c) + [1-\alpha(x_t)]v^{-}(x_t, t, c).
\end{equation}

Since $\pi^+(x_0\mid c)=p^{old}(x_0\mid O{=}1,c)=p^{\ft}(x_0\mid c)$ by definition, we have $\pi^+_t=p^{\ft}_t$, and from equation~\ref{p_ft},
\[
\frac{p_t^{\ft}(x_t\mid c)}{p_t^{old}(x_t\mid c)}=\frac{h(x_t,t,c)}{Z(c)}.
\]
Since $\E_{\pi^{old}(x_0\mid c)}[w(x_0)]=Z(c)$, substituting gives
\[
\alpha(x_t)=\frac{h(x_t,t,c)}{Z(c)}\cdot Z(c)=h(x_t,t,c).
\]

As noted above, $\pi^+_t\equiv p_t^{\ft}$ as distributions, hence their posteriors over $x_0$ given $x_t$ coincide, and since the velocity field is a linear functional of the posterior mean,
\[
v^+(x_t,t,c)=v^{\ft}(x_t,t,c)=v^{\base}(x_t,t,c)-\frac{t}{1-t}\nabla_x\log h(x_t,t,c).
\]

\subsection{The Guidance Direction {$\Delta$} and {$v^-$} as Anti-Guidance}

According to equation~\ref{eq:velocity_split}, $v^{old}, v^+, v^-$ are on the same line. DiffuionNFT defines velocity improvement direction $\Delta$ as:
\[
\Delta:=\alpha(x_t)\,[v^+-v^{old}]=[1-\alpha(x_t)]\,[v^{old}-v^-].
\]
Using $\alpha=h$ and $v^+-v^{old}=v^{\ft}-v^{old}=-\frac{t}{1-t}\nabla_x\log h$,
\[
\Delta=h\cdot\Big(-\frac{t}{1-t}\nabla_x\log h\Big)=-\frac{t}{1-t}\,h\,\nabla_x\log h=-\frac{t}{1-t}\nabla_x h,
\]
using $h\nabla\log h=\nabla h$. That is, $\Delta$ is the (unnormalized) gradient $\nabla_x h$, rescaled by $-\frac{t}{1-t}$.

For $v^-$: by the same construction with $w^-(x_0):=1-w(x_0)$, $\pi^-=p^{old}(\cdot\mid O{=}0,c)$ has $h^-(x_t,t,c)=1-h(x_t,t,c)$, and
\[
v^-(x_t,c,t)=v^{old}(x_t,t,c)-\frac{t}{1-t}\nabla_x\log\big(1-h(x_t,t,c)\big).
\]
Substituting into the first form of $\Delta$, with $\nabla_x\log(1-h)=-\nabla_x h/(1-h)$,
\[
-[1-h]\Big[v^{old}-v^-\Big]=-[1-h]\cdot\frac{t}{1-t}\cdot\frac{\nabla_x h}{1-h}=-\frac{t}{1-t}\nabla_x h,
\]
confirming consistency. Thus $v^-$ is precisely the $h$-transform velocity for the \emph{complementary} event $\{O=0\}$: it is not an independent construction but an ``anti-guidance'' field pointing away from the optimal region.

\subsection{Double-Branch Optimal Velocity}

DiffusionNFT trains $v_\theta$ (it acts as an anchor for improvement guidance while training, and used for inference after training) via
\[
\mathcal L(\theta)=\E_{c,\pi^{old}(x_0\mid c),t}\Big[w(x_0)\,\|v_\theta^+-v\|_2^2+(1-w(x_0))\,\|v_\theta^--v\|_2^2\Big],
\]
where
\begin{equation}\label{def:v_param}
    v_\theta^+:=(1-\beta)v^{old}+\beta v_\theta, \quad v_\theta^-:=(1+\beta)v^{old}-\beta v_\theta.
\end{equation}

The 'good velocity' is parametrized as a linear combination between anchor $v_\theta$ and $v^{old}$, and 'bad velocity' is pushing $v^{old}$ to the opposite direction from the 'good' one.

By definition $p^{\ft}(x_0\mid c)=\frac{w(x_0)}{Z(c)}p^{old}(x_0\mid c)$, i.e.
\[
w(x_0)\,p^{old}(x_0\mid c)=Z(c)\,p^{\ft}(x_0\mid c).
\]
Multiplying both sides by the forward transition kernel $\phi(x_t\mid x_0)$, which is shared by $p^{old}$ and $p^{\ft}$, and applying Bayes' rule $p(x_0\mid c)\phi(x_t\mid x_0)=p_t(x_t\mid c)\,p_{0|t}(x_0\mid x_t,c)$ on each side:
\[
w(x_0)\,p_t^{old}(x_t\mid c)\,\pi^{old}_{0|t}(x_0\mid x_t,c)=Z(c)\,p_t^{\ft}(x_t\mid c)\,\pi^+_{0|t}(x_0\mid x_t,c).
\]
Dividing both sides by $p_t^{old}(x_t\mid c)$ and using $\dfrac{p_t^{\ft}(x_t\mid c)}{p_t^{old}(x_t\mid c)}=\dfrac{h(x_t,t,c)}{Z(c)}$ in Equation~\ref{p_ft}\,
\begin{equation}
\label{eq:old_pos_relation}
  w(x_0)\,\pi^{old}_{0|t}(x_0\mid x_t,c)=Z(c)\cdot\frac{h(x_t,t,c)}{Z(c)}\cdot\pi^+_{0|t}(x_0\mid x_t,c)=h(x_t,t,c)\,\pi^+_{0|t}(x_0\mid x_t,c).  
\end{equation}

The same argument with $w$ replaced by $1-w$ (equivalently $h$ replaced by $1-h$, $\pi^+$ by $\pi^-$) gives $(1-w(x_0))\,\pi^{old}_{0|t}=(1-h(x_t,t,c))\,\pi^-_{0|t}$.

\paragraph{Reduction to a weighted regression at fixed $x_t$.}
In NFT, loss function is defined as 
\[
\mathcal{L}(\theta, c) = \mathbb{E}_{\pi^{old}(x_0,x_t \mid c)}[w(x_0)\|v_{\theta}^+(x_t,t,c) - v\|_2^2 + (1-w(x_0))\|v_{\theta}^-(x_t,t,c) - v\|_2^2]
\]
Since the joint density $\pi^{old}(x_0,x_t \mid c)$ factorizes as $\pi_t^{old}(x_t\mid c)\,\pi^{old}_{0|t}(x_0\mid x_t,c)$, the inner expectation of the positive term at fixed $(x_t,t,c)$ is
\[
\int w(x_0)\,\pi^{old}_{0|t}(x_0\mid x_t,c)\,\|v_\theta^+-v\|^2\,dx_0
=h(x_t,t,c)\int\pi^+_{0|t}(x_0\mid x_t,c)\,\|v_\theta^+-v\|^2\,dx_0
=h\,\E_{\pi^+_{0|t}}\big[\|v_\theta^+-v\|^2\big],
\]
using the identity~\ref{eq:old_pos_relation} and $v_\theta^+$ being independent of $x_0$. Since $v_\theta^+$ is a constant w.r.t.\ $x_0$,
\[
\E_{\pi^+_{0|t}}\big[\|v_\theta^+-v\|^2\big]=\big\|v_\theta^+-\E_{\pi^+_{0|t}}[v]\big\|^2+\underbrace{Var_{\pi^+_{0|t}}(v)}_{\text{independent of }\theta}
=\|v_\theta^+-v^+\|^2+C_+,
\]
where $\E_{\pi^+_{0|t}}[v]=v^+(x_t,c,t)$ by definition of the velocity field. The negative term reduces identically to $(1-h)\big[\|v_\theta^--v^-\|^2+C_-\big]$. Dropping the $\theta$-independent constants, the loss at fixed $x_t,t,c$ is equivalent to minimizing
\[
h\,\|v_\theta^+-v^+\|_2^2+(1-h)\,\|v_\theta^--v^-\|_2^2.
\]

Let $Y:=v_\theta-v^{old}$.
Let $D:=v^{+}-v^{old}$ and $D^-:=v^--v^{old}$. Substituting the parameterization,
\[
v_\theta^+-v^+=(1-\beta)v^{old}+\beta v_\theta-(v^{old}+D)=\beta(v_\theta-v^{old})-D=\beta\Big(Y-\tfrac{D}{\beta}\Big),
\]
\[
v_\theta^--v^-=(1+\beta)v^{old}-\beta v_\theta-(v^{old}+D^-)=-\beta(v_\theta-v^{old})-D^-=-\beta\Big(Y+\tfrac{D^-}{\beta}\Big).
\]
Hence $\|v_\theta^+-v^+\|^2=\beta^2\|Y-D/\beta\|^2$ and $\|v_\theta^--v^-\|^2=\beta^2\|Y+D^-/\beta\|^2$; factoring out $\beta^2>0$ (which does not affect the minimizer) gives the weighted least-squares problem
\[
f(Y)=h\Big\|Y-\tfrac{D}{\beta}\Big\|_2^2+(1-h)\Big\|Y+\tfrac{D^-}{\beta}\Big\|_2^2.
\]

\paragraph{Solving for the minimizer.}
$f$ is a weighted sum of squared distances to two points, $D/\beta$ (weight $h$) and $-D^-/\beta$ (weight $1-h$); its minimizer is the weighted average
\[
Y^*=h\cdot\frac{D}{\beta}+(1-h)\cdot\Big(-\frac{D^-}{\beta}\Big)=\frac{hD-(1-h)D^-}{\beta}.
\]
Since $v^{old}=h\,v^++(1-h)\,v^-$, subtracting $v^{old}$ from both sides gives $0=hD+(1-h)D^-$, i.e.\ $hD=-(1-h)D^-$. Substituting,
\[
Y^*=\frac{hD-(1-h)D^-}{\beta}=\frac{hD+hD}{\beta}=\frac{2hD}{\beta}
\]
\[\Longrightarrow\quad
v_{\theta^*}=v^{old}-\frac{2h(x_t,t,c)}{\beta}\cdot\frac{t}{1-t}\nabla_x\log h(x_t,t,c).
\]
For fixed scalar $\beta$, this is a state-dependent damped version of $v^{\ft}$: the guidance strength scales with the local optimality probability $h$, vanishing where $h\to0$.

Therefore our algorithm is choosing $\beta$ in the Definition~\ref{def:v_param} to be $2h(x_t,t,c)$. This is reasonable because when you have $(x_t,t \mid c)$ is more likely to be optimal, that is $h(x_t,t,c) > 0.5$, we are more confident to push fine-tuned sampling strategy $v_\theta$ to be farther on the current improvement direction. On the other hand if $h(x_t,t,c) < 0.5$, we are not willing to trust $v_\theta$ that much and keep closer to $v^{old}$. However, in DiffusionNFT, they don't estimate $h$-function explicitly, so choosing $\beta$ as a function depending on $(x_t,t,c)$ is not viable. They have to leave $\beta$ as a hyper-parameter (in their implementation, $\beta=1$ is default value). Our algorithm provided more flexibility on this aspect.

\subsection{Single-Branch Optimal Velocity}

Training on the positive branch alone, $\mathcal L^+(\theta)=\E_{c,t,\pi^{old}(x_0\mid c)}\big[w\,\|v_\theta^+-v\|_2^2\big]$, and using $w\,\pi^{old}_{0|t}=h\,\pi^+_{0|t}$,
\[
\mathcal L^+(\theta)=\E_{c,t,x_t}\Big[h(x_t,t,c)\cdot\E_{x_0\sim\pi^+_{0|t}}\|v_\theta^+-v\|_2^2\Big].
\]
Pointwise in $x_t$, the $L^2$-minimizer is the conditional mean regardless of the (positive) weight $h$:
\[
v_\theta^{+*}=\E_{\pi^+_{0|t}}[v]=v^+=v^{\ft}
\quad\Longrightarrow\quad
v_\theta^*=v^{\base}+\frac{D}{\beta}=v^{\base}-\frac{1}{\beta}\cdot\frac{t}{1-t}\nabla_x\log h,
\]
which is \emph{exact}, with no $h$-dependent damping, for any fixed $\beta$.

\subsection{Unifying the Three via {$h$}}

Comparing the two population optima:
\[
\text{double-branch: } v_{\theta^*}=v^{\base}-\frac{2h}{\beta}\cdot\frac{t}{1-t}\nabla\log h,
\qquad
\text{single-branch: } v_{\theta^*}=v^{\base}-\frac{1}{\beta}\cdot\frac{t}{1-t}\nabla\log h.
\]
Setting $\beta(x_t,t,c)=2h(x_t,t,c)$ in the double-branch objective exactly cancels the damping factor, giving $v_{\theta^*}=v^{\ft}$; equivalently, this coincides with the single-branch optimum at $\beta=1$. Thus
\[
\text{double-branch},\ \beta=2h
\quad\Longleftrightarrow\quad
\text{single-branch},\ \beta=1
\quad\Longleftrightarrow\quad
v_\theta^*=v^{\ft}=v^{\base}-\frac{t}{1-t}\nabla_x\log h.
\]

\section{Discussion on Contingency Table}

\subsection{Birthday-Deathday Data}
The following Table~\ref{tab:birthday_deathday} is taken from the classical contingency-table example of
\cite{diaconis1998algebraic}.
\begin{table}[t]
\centering
\small
\setlength{\tabcolsep}{2.2pt}
\begin{tabular}{c|rrrrrrrrrrrr|r}
\toprule
Birth $\backslash$ Death
& Jan & Feb & Mar & Apr & May & Jun
& Jul & Aug & Sep & Oct & Nov & Dec & Total\\
\midrule
Jan & 1&0&0&0&1&2&0&0&1&0&1&0&6\\
Feb & 1&0&0&1&0&0&0&0&0&1&0&2&5\\
Mar & 1&0&0&0&2&1&0&0&0&0&0&1&5\\
Apr & 3&0&2&0&0&0&1&0&1&3&1&1&12\\
May & 2&1&1&1&1&1&1&1&1&1&1&0&12\\
Jun & 2&0&0&0&1&0&0&0&0&0&0&0&3\\
Jul & 2&0&2&1&0&0&0&0&1&1&1&2&10\\
Aug & 0&0&0&3&0&0&1&0&0&1&0&2&7\\
Sep & 0&0&0&1&1&0&0&0&0&0&1&0&3\\
Oct & 1&1&0&2&0&0&1&0&0&1&1&0&7\\
Nov & 0&1&1&1&2&0&0&2&0&1&1&0&9\\
Dec & 0&1&1&0&0&0&1&0&0&0&0&0&3\\
\midrule
Total
&13&4&7&10&8&4&5&3&4&9&7&8&82\\
\bottomrule
\end{tabular}
\caption{Birthday and death month of 82 descendants of Queen Victoria.}
\label{tab:birthday_deathday}
\end{table}

\subsection{Pretraining the Contingency-Table Flow Model}
\label{contingency_pretrained}

To obtain the pretrained generative model used in the contingency-table
experiment, we first construct a training distribution over $12\times12$
nonnegative tables with a fixed total count $N=82$.
For each training sample, we randomly generate an integer table
$X\in\mathbb{Z}_{+}^{12\times12}$ satisfying
\[
\sum_{i,j} X_{ij}=N.
\]
We then normalize the table by its total count,
\[
P_{ij}=\frac{X_{ij}}{N},
\qquad
\sum_{i,j}P_{ij}=1,
\]
so that each table can be viewed as a probability vector on the
$144$-dimensional simplex. A continuous flow model is pretrained on
these normalized probability tables.

At inference time, the generated probability table
$\widehat P$ is converted back to an integer contingency table with
total count exactly $N$. We first compute
\[
Y_{ij}=N\widehat P_{ij},
\qquad
\widetilde X_{ij}=\lfloor Y_{ij}\rfloor .
\]
Since rounding down may leave a residual
\[
R=N-\sum_{i,j}\widetilde X_{ij},
\]
we assign one additional count to the $R$ entries having the largest
fractional parts
\[
Y_{ij}-\lfloor Y_{ij}\rfloor.
\]
This largest-remainder rounding procedure guarantees
\[
\sum_{i,j}\widetilde X_{ij}=N
\]
while keeping the resulting integer table as close as possible to the
continuous model output under this rounding scheme.
If the total number is not hard-constrained to be the total number, we could let 
\[
Y_{ij}=N\widehat P_{ij},
\qquad
\widetilde X_{ij}=\textbf{round($Y_{ij}$)}.
\]

\section{Additional Details on Text-to-Image Task}
\label{appdx_t2i_exp}
\paragraph{ODE Integration.} Roll-out time steps are $40$ for both training and inference. Note that we only solve the flow ODE step-by-step till $\nabla \log h$, then use one step estimation no matter using Estimator~\ref{est:MC} or \ref{est:fd}. However, the solver used in training roll-outs is Euler, and during inference is DPM-Solver$++$\cite{Lu_2025}, and CFG scale for online $h$-guidance w/ CFG is $4$ by default, for SD3.5-M is $3.5$ by default (both of them are taken due to the relatively best performance among different metrics).

\paragraph{Online Data Collection.} Slow mixed policy is used to generate online samples. Follow the default design used in DiffusionNFT\cite{zheng2026diffusionNFT}, $\eta_k = \min(0.001 \cdot k, 0.5)$, which means the data collection starts from on-policy, and gradually converges to $0.5 \cdot v^{old} + 0.5 \cdot v_{\theta}$ after 500 iterations. On the other hand, for the off policy guidance baseline, the roll-out policy and regression target are based on the base model $v^{\base}$.

\paragraph{Other Configurations.} In our T2I implementation, 7 Nvidia L40S is used. For each iteration per GPU, 4 prompts are randomly picked, and 12 initial noise is generated for each prompt. We do not use all time steps in training because two denominators in Flow-equivalent SDE~\ref{eq:flow_sde} will approach to zero when $t \rightarrow 1$, therefore we set $t_{\min}=0.05$ and $t_{\max}=0.9$. Moreover, stochasticity parameter $\lambda_t=1$ in SDE~\ref{eq:flow_sde}. Learning Rate is $3e-5$, Gradient Clip is set to be $1.0$, and slight LoRA regularization $1e-5$ is applied.

\section{Additional Qualitative Comparison}
In this section, we will provide more images showing the aesthetic quality and semantic alignment provided by our algorithm. For baselines, SD3.5-Medium, DiffusionNFT, FlowGRPO and off-policy h-guidance are compared in images.

\begin{figure*}[t]
    \centering
    \setlength{\tabcolsep}{5pt}
    \renewcommand{\arraystretch}{1.0}

    \begin{tabular}{>{\centering\arraybackslash}m{0.17\textwidth} c c c c}
        & \textbf{Prompt 1} & \textbf{Prompt 2} & \textbf{Prompt 3} & \textbf{Prompt 4} \\[2mm]

        \shortstack{\textbf{SD3.5M}\\\textbf{(w/o CFG)}}
        & \includegraphics[width=0.16\textwidth,valign=c]{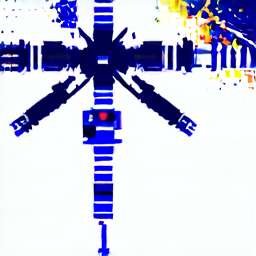}
        & \includegraphics[width=0.16\textwidth,valign=c]{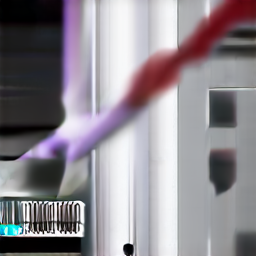}
        & \includegraphics[width=0.16\textwidth,valign=c]{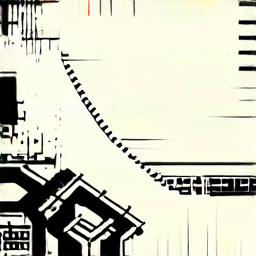}
        & \includegraphics[width=0.16\textwidth,valign=c]{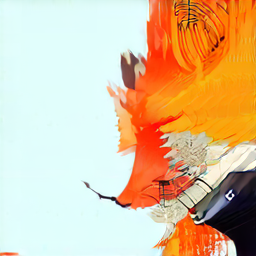}\\
        \noalign{\vskip 3mm}
        
        \shortstack{\textbf{SD3.5M}\\\textbf{(w/ CFG)}}
        & \includegraphics[width=0.16\textwidth,valign=c]{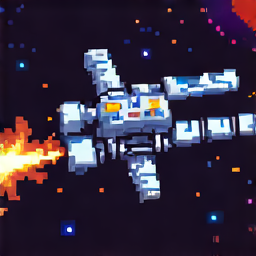}
        & \includegraphics[width=0.16\textwidth,valign=c]{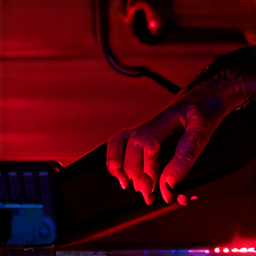}
        & \includegraphics[width=0.16\textwidth,valign=c]{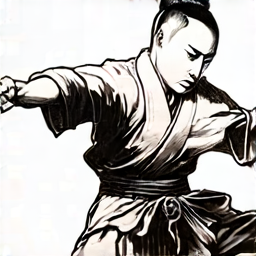}
        & \includegraphics[width=0.16\textwidth,valign=c]{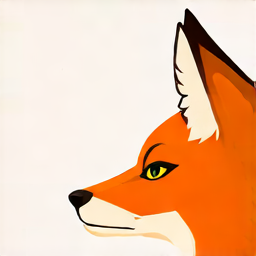}\\
        \noalign{\vskip 3mm}

        \textbf{DiffusionNFT}
        & \includegraphics[width=0.16\textwidth,valign=c]{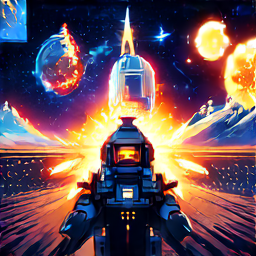}
        & \includegraphics[width=0.16\textwidth,valign=c]{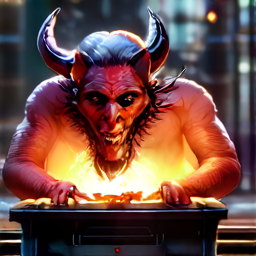}
        & \includegraphics[width=0.16\textwidth,valign=c]{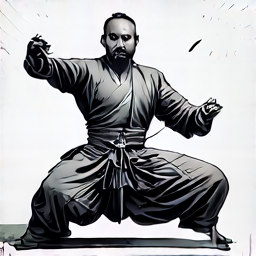}
        & \includegraphics[width=0.16\textwidth,valign=c]{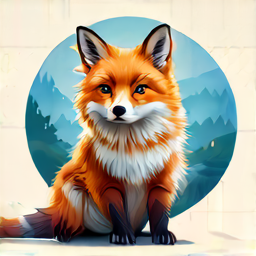} \\\noalign{\vskip 3mm}

        \textbf{FlowGRPO}
        & \includegraphics[width=0.16\textwidth,valign=c]{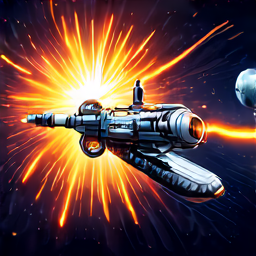}
        & \includegraphics[width=0.16\textwidth,valign=c]{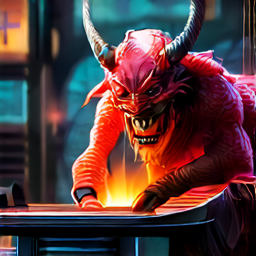}
        & \includegraphics[width=0.16\textwidth,valign=c]{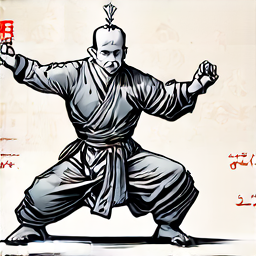}
        & \includegraphics[width=0.16\textwidth,valign=c]{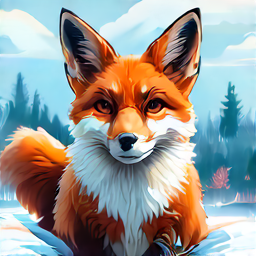} \\\noalign{\vskip 3mm}

        \shortstack{\textbf{Off-policy}\\\textbf{$h$-guidance}}
        & \includegraphics[width=0.16\textwidth,valign=c]{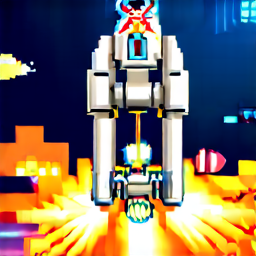}
        & \includegraphics[width=0.16\textwidth,valign=c]{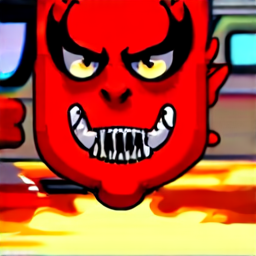}
        & \includegraphics[width=0.16\textwidth,valign=c]{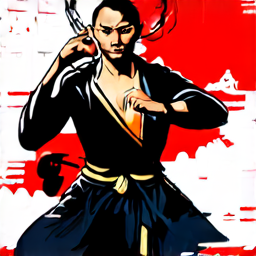}
        & \includegraphics[width=0.16\textwidth,valign=c]{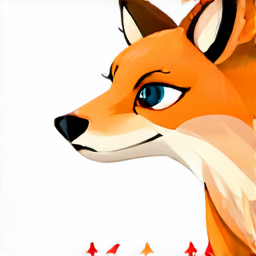} \\\noalign{\vskip 3mm}

        \shortstack{\textbf{DOHF}\\\textbf{(w/o CFG)}}
        & \includegraphics[width=0.16\textwidth,valign=c]{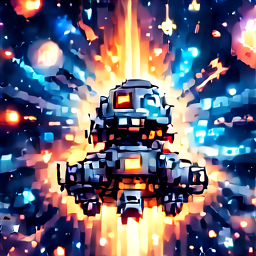}
        & \includegraphics[width=0.16\textwidth,valign=c]{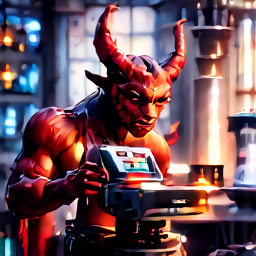}
        & \includegraphics[width=0.16\textwidth,valign=c]{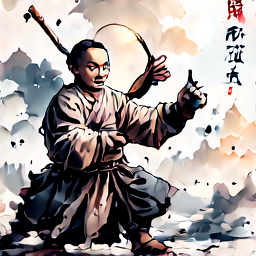}
        & \includegraphics[width=0.16\textwidth,valign=c]{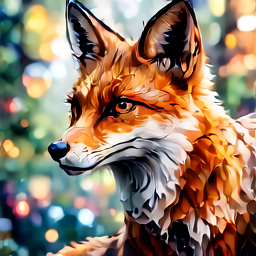} \\\noalign{\vskip 3mm}

        \shortstack{\textbf{DOHF}\\\textbf{(w/ CFG)}}
        & \includegraphics[width=0.16\textwidth,valign=c]{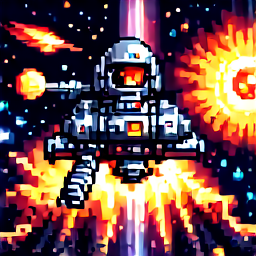}
        & \includegraphics[width=0.16\textwidth,valign=c]{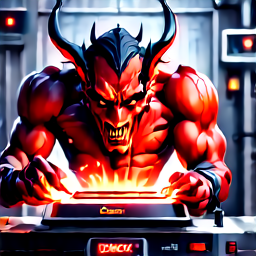}
        & \includegraphics[width=0.16\textwidth,valign=c]{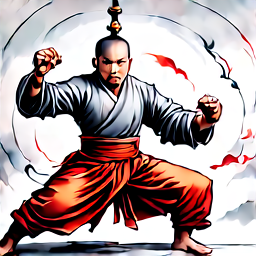}
        & \includegraphics[width=0.16\textwidth,valign=c]{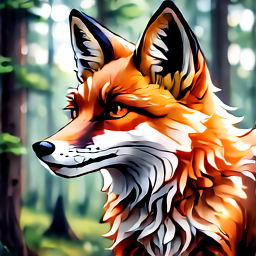} 
    \end{tabular}

    \caption{Qualitative comparison across different methods and prompts. The prompts are:
    (Prompt 1) ``Pixel art of a space station launching with two suns and a comet, featuring a retro space helmet, minigun, and a bright explosion in a dream-like atmosphere with CGA shading.";
    (Prompt 2) ``The devil using a heat press.";
    (Prompt 3) ``Ink drawing of a Shaolin monk in fighting pose.";
    (Prompt 4) ``A side view of a fox, with a flat design, Artwork of t-shirt graphic."}
    \label{fig:additional_aesthetic}
\end{figure*}

\begin{figure*}[t]
    \centering
    \setlength{\tabcolsep}{5pt}
    \renewcommand{\arraystretch}{1.0}

    \begin{tabular}{>{\centering\arraybackslash}m{0.17\textwidth} c c c c}
        & \textbf{Prompt 1} & \textbf{Prompt 2} & \textbf{Prompt 3} & \textbf{Prompt 4} \\[2mm]

        \shortstack{\textbf{SD3.5M}\\\textbf{(w/o CFG)}}
        & \includegraphics[width=0.16\textwidth,valign=c]{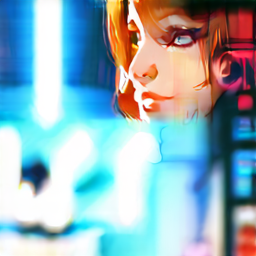}
        & \includegraphics[width=0.16\textwidth,valign=c]{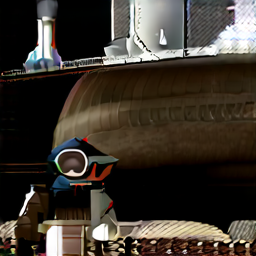}
        & \includegraphics[width=0.16\textwidth,valign=c]{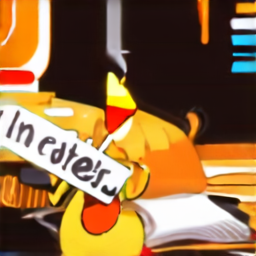}
        & \includegraphics[width=0.16\textwidth,valign=c]{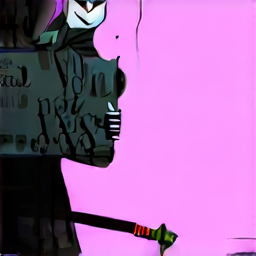}\\
        \noalign{\vskip 3mm}

        \shortstack{\textbf{SD3.5M}\\\textbf{(w/ CFG)}}
        & \includegraphics[width=0.16\textwidth,valign=c]{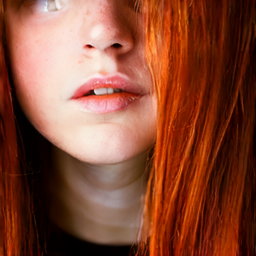}
        & \includegraphics[width=0.16\textwidth,valign=c]{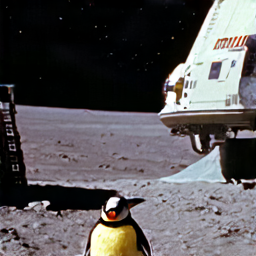}
        & \includegraphics[width=0.16\textwidth,valign=c]{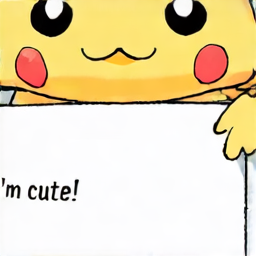}
        & \includegraphics[width=0.16\textwidth,valign=c]{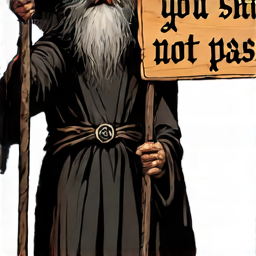}\\
        \noalign{\vskip 3mm}

        \textbf{DiffusionNFT}
        & \includegraphics[width=0.16\textwidth,valign=c]{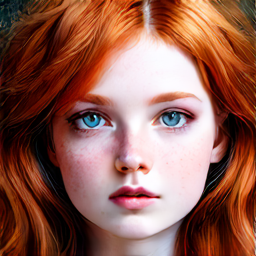}
        & \includegraphics[width=0.16\textwidth,valign=c]{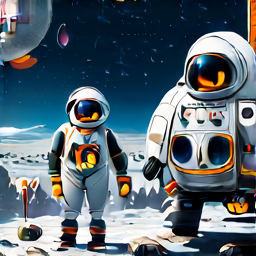}
        & \includegraphics[width=0.16\textwidth,valign=c]{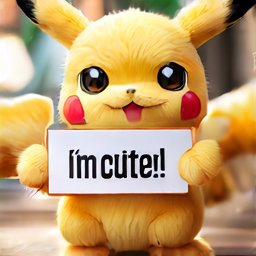}
        & \includegraphics[width=0.16\textwidth,valign=c]{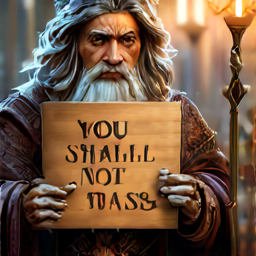} \\\noalign{\vskip 3mm}

        \textbf{FlowGRPO}
        & \includegraphics[width=0.16\textwidth,valign=c]{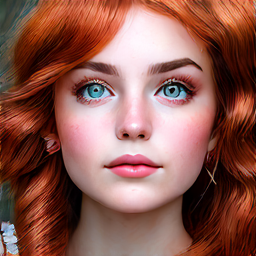}
        & \includegraphics[width=0.16\textwidth,valign=c]{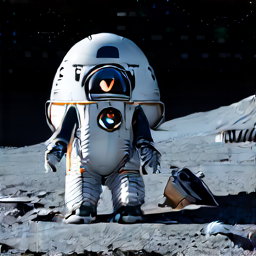}
        & \includegraphics[width=0.16\textwidth,valign=c]{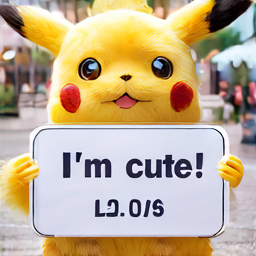}
        & \includegraphics[width=0.16\textwidth,valign=c]{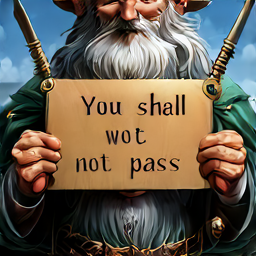} \\\noalign{\vskip 3mm}

        \shortstack{\textbf{Off-policy}\\\textbf{$h$-guidance}}
        & \includegraphics[width=0.16\textwidth,valign=c]{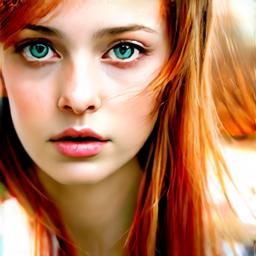}
        & \includegraphics[width=0.16\textwidth,valign=c]{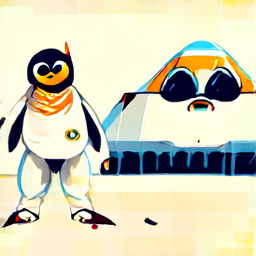}
        & \includegraphics[width=0.16\textwidth,valign=c]{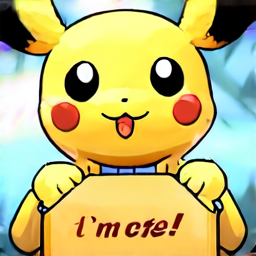}
        & \includegraphics[width=0.16\textwidth,valign=c]{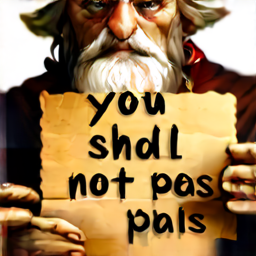} \\\noalign{\vskip 3mm}

        \shortstack{\textbf{DOHF}\\\textbf{(w/o CFG)}}
        & \includegraphics[width=0.16\textwidth,valign=c]{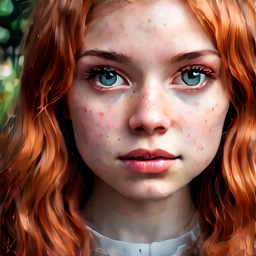}
        & \includegraphics[width=0.16\textwidth,valign=c]{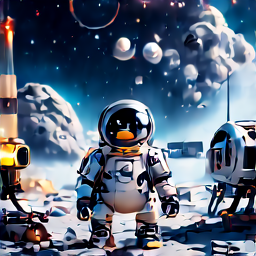}
        & \includegraphics[width=0.16\textwidth,valign=c]{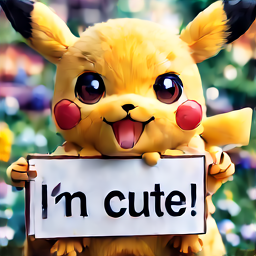}
        & \includegraphics[width=0.16\textwidth,valign=c]{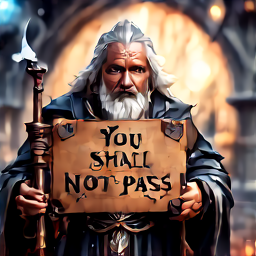} \\\noalign{\vskip 3mm}

        \shortstack{\textbf{DOHF}\\\textbf{(w/ CFG)}}
        & \includegraphics[width=0.16\textwidth,valign=c]{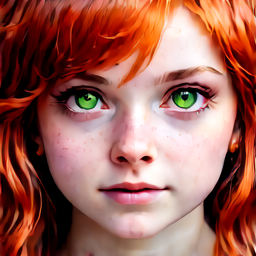}
        & \includegraphics[width=0.16\textwidth,valign=c]{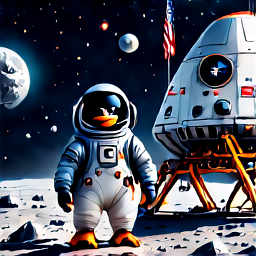}
        & \includegraphics[width=0.16\textwidth,valign=c]{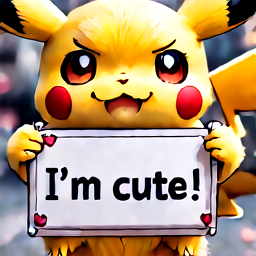}
        & \includegraphics[width=0.16\textwidth,valign=c]{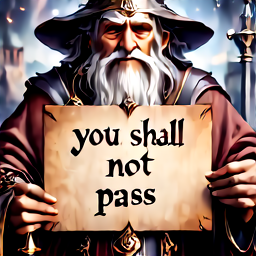} 
    \end{tabular}

    \caption{Qualitative comparison across different methods and prompts. The prompts are:
    (Prompt 1, color) ``A young woman with \textcolor{red}{red} hairs and \textcolor{green}{green} eyes";
    (Prompt 2, spatial) ``A penguin astronaut, in a penguin spacesuit, standing \underline{next} to the apollo landing";
    (Prompt 3, text) ``A pikachu holding a sign that says \textbf{"I'm cute!"}";
    (Prompt 4, text) ``Gandalf holding a sign with a text \textbf{"you shall not pass"}".}
    \label{fig:additional_semantic}
\end{figure*}

For aesthetic quality, refer to Figure~\ref{fig:additional_aesthetic}. For semantic alignment, refer to Figure~\ref{fig:additional_semantic}.

\end{document}